\documentclass[letterpaper,12pt,english]{article}
\usepackage[utf8]{inputenc}
\DeclareUnicodeCharacter{00A0}{\nobreakspace}
\usepackage{cmap}
\usepackage[T1]{fontenc}
\usepackage{babel}
\usepackage{times}
\usepackage[Bjarne]{fncychap}
\usepackage{longtable}
\usepackage{sphinx}
\usepackage{multirow}
\usepackage{eqparbox}
\usepackage{amsfonts}

\addto\captionsenglish{}
\addto\captionsenglish{}
\SetupFloatingEnvironment{literal-block}{name=Listing }

  \pdfoutput=1
  \usepackage{cmap}
  \usepackage{amssymb}
  \newcommand{\chapter}[1]{}  % hack to be able to use article documentclass
  \newcommand{\ignore}[1]{}  % never used

\title{{COCO}: Performance Assessment}
\date{May 11, 2016}
\release{0.9}
\date{\vspace{-1ex}}\author{Nikolaus Hansen$^{1,2}$, 
      Anne Auger$^{1,2}$, 
      Dimo Brockhoff$^3$,
      Dejan Tu\v{s}ar$^3$, 
      Tea Tu\v{s}ar$^3$
  \\
    $^1$Inria, research centre Saclay, France
  \\
   $^2$Universit\'e Paris-Saclay, LRI, France
  \\
    $^3$Inria, research centre Lille, France
    }

\renewcommand{\releasename}{Release}

\makeindex

\makeatletter
\def\PYG@reset{\let\PYG@it=\relax \let\PYG@bf=\relax%
    \let\PYG@ul=\relax \let\PYG@tc=\relax%
    \let\PYG@bc=\relax \let\PYG@ff=\relax}
\def\PYG@tok#1{\csname PYG@tok@#1\endcsname}
\def\PYG@toks#1+{\ifx\relax#1\empty\else%
    \PYG@tok{#1}\expandafter\PYG@toks\fi}
\def\PYG@do#1{\PYG@bc{\PYG@tc{\PYG@ul{%
    \PYG@it{\PYG@bf{\PYG@ff{#1}}}}}}}
\def\PYG#1#2{\PYG@reset\PYG@toks#1+\relax+\PYG@do{#2}}

\expandafter\def\csname PYG@tok@gd\endcsname{\def\PYG@tc##1{\textcolor[rgb]{0.63,0.00,0.00}{##1}}}
\expandafter\def\csname PYG@tok@gu\endcsname{\let\PYG@bf=\textbf\def\PYG@tc##1{\textcolor[rgb]{0.50,0.00,0.50}{##1}}}
\expandafter\def\csname PYG@tok@gt\endcsname{\def\PYG@tc##1{\textcolor[rgb]{0.00,0.27,0.87}{##1}}}
\expandafter\def\csname PYG@tok@gs\endcsname{\let\PYG@bf=\textbf}
\expandafter\def\csname PYG@tok@gr\endcsname{\def\PYG@tc##1{\textcolor[rgb]{1.00,0.00,0.00}{##1}}}
\expandafter\def\csname PYG@tok@cm\endcsname{\let\PYG@it=\textit\def\PYG@tc##1{\textcolor[rgb]{0.25,0.50,0.56}{##1}}}
\expandafter\def\csname PYG@tok@vg\endcsname{\def\PYG@tc##1{\textcolor[rgb]{0.73,0.38,0.84}{##1}}}
\expandafter\def\csname PYG@tok@vi\endcsname{\def\PYG@tc##1{\textcolor[rgb]{0.73,0.38,0.84}{##1}}}
\expandafter\def\csname PYG@tok@mh\endcsname{\def\PYG@tc##1{\textcolor[rgb]{0.13,0.50,0.31}{##1}}}
\expandafter\def\csname PYG@tok@cs\endcsname{\def\PYG@tc##1{\textcolor[rgb]{0.25,0.50,0.56}{##1}}\def\PYG@bc##1{\setlength{\fboxsep}{0pt}\colorbox[rgb]{1.00,0.94,0.94}{\strut ##1}}}
\expandafter\def\csname PYG@tok@ge\endcsname{\let\PYG@it=\textit}
\expandafter\def\csname PYG@tok@vc\endcsname{\def\PYG@tc##1{\textcolor[rgb]{0.73,0.38,0.84}{##1}}}
\expandafter\def\csname PYG@tok@il\endcsname{\def\PYG@tc##1{\textcolor[rgb]{0.13,0.50,0.31}{##1}}}
\expandafter\def\csname PYG@tok@go\endcsname{\def\PYG@tc##1{\textcolor[rgb]{0.20,0.20,0.20}{##1}}}
\expandafter\def\csname PYG@tok@cp\endcsname{\def\PYG@tc##1{\textcolor[rgb]{0.00,0.44,0.13}{##1}}}
\expandafter\def\csname PYG@tok@gi\endcsname{\def\PYG@tc##1{\textcolor[rgb]{0.00,0.63,0.00}{##1}}}
\expandafter\def\csname PYG@tok@gh\endcsname{\let\PYG@bf=\textbf\def\PYG@tc##1{\textcolor[rgb]{0.00,0.00,0.50}{##1}}}
\expandafter\def\csname PYG@tok@ni\endcsname{\let\PYG@bf=\textbf\def\PYG@tc##1{\textcolor[rgb]{0.84,0.33,0.22}{##1}}}
\expandafter\def\csname PYG@tok@nl\endcsname{\let\PYG@bf=\textbf\def\PYG@tc##1{\textcolor[rgb]{0.00,0.13,0.44}{##1}}}
\expandafter\def\csname PYG@tok@nn\endcsname{\let\PYG@bf=\textbf\def\PYG@tc##1{\textcolor[rgb]{0.05,0.52,0.71}{##1}}}
\expandafter\def\csname PYG@tok@no\endcsname{\def\PYG@tc##1{\textcolor[rgb]{0.38,0.68,0.84}{##1}}}
\expandafter\def\csname PYG@tok@na\endcsname{\def\PYG@tc##1{\textcolor[rgb]{0.25,0.44,0.63}{##1}}}
\expandafter\def\csname PYG@tok@nb\endcsname{\def\PYG@tc##1{\textcolor[rgb]{0.00,0.44,0.13}{##1}}}
\expandafter\def\csname PYG@tok@nc\endcsname{\let\PYG@bf=\textbf\def\PYG@tc##1{\textcolor[rgb]{0.05,0.52,0.71}{##1}}}
\expandafter\def\csname PYG@tok@nd\endcsname{\let\PYG@bf=\textbf\def\PYG@tc##1{\textcolor[rgb]{0.33,0.33,0.33}{##1}}}
\expandafter\def\csname PYG@tok@ne\endcsname{\def\PYG@tc##1{\textcolor[rgb]{0.00,0.44,0.13}{##1}}}
\expandafter\def\csname PYG@tok@nf\endcsname{\def\PYG@tc##1{\textcolor[rgb]{0.02,0.16,0.49}{##1}}}
\expandafter\def\csname PYG@tok@si\endcsname{\let\PYG@it=\textit\def\PYG@tc##1{\textcolor[rgb]{0.44,0.63,0.82}{##1}}}
\expandafter\def\csname PYG@tok@s2\endcsname{\def\PYG@tc##1{\textcolor[rgb]{0.25,0.44,0.63}{##1}}}
\expandafter\def\csname PYG@tok@nt\endcsname{\let\PYG@bf=\textbf\def\PYG@tc##1{\textcolor[rgb]{0.02,0.16,0.45}{##1}}}
\expandafter\def\csname PYG@tok@nv\endcsname{\def\PYG@tc##1{\textcolor[rgb]{0.73,0.38,0.84}{##1}}}
\expandafter\def\csname PYG@tok@s1\endcsname{\def\PYG@tc##1{\textcolor[rgb]{0.25,0.44,0.63}{##1}}}
\expandafter\def\csname PYG@tok@ch\endcsname{\let\PYG@it=\textit\def\PYG@tc##1{\textcolor[rgb]{0.25,0.50,0.56}{##1}}}
\expandafter\def\csname PYG@tok@m\endcsname{\def\PYG@tc##1{\textcolor[rgb]{0.13,0.50,0.31}{##1}}}
\expandafter\def\csname PYG@tok@gp\endcsname{\let\PYG@bf=\textbf\def\PYG@tc##1{\textcolor[rgb]{0.78,0.36,0.04}{##1}}}
\expandafter\def\csname PYG@tok@sh\endcsname{\def\PYG@tc##1{\textcolor[rgb]{0.25,0.44,0.63}{##1}}}
\expandafter\def\csname PYG@tok@ow\endcsname{\let\PYG@bf=\textbf\def\PYG@tc##1{\textcolor[rgb]{0.00,0.44,0.13}{##1}}}
\expandafter\def\csname PYG@tok@sx\endcsname{\def\PYG@tc##1{\textcolor[rgb]{0.78,0.36,0.04}{##1}}}
\expandafter\def\csname PYG@tok@bp\endcsname{\def\PYG@tc##1{\textcolor[rgb]{0.00,0.44,0.13}{##1}}}
\expandafter\def\csname PYG@tok@c1\endcsname{\let\PYG@it=\textit\def\PYG@tc##1{\textcolor[rgb]{0.25,0.50,0.56}{##1}}}
\expandafter\def\csname PYG@tok@o\endcsname{\def\PYG@tc##1{\textcolor[rgb]{0.40,0.40,0.40}{##1}}}
\expandafter\def\csname PYG@tok@kc\endcsname{\let\PYG@bf=\textbf\def\PYG@tc##1{\textcolor[rgb]{0.00,0.44,0.13}{##1}}}
\expandafter\def\csname PYG@tok@c\endcsname{\let\PYG@it=\textit\def\PYG@tc##1{\textcolor[rgb]{0.25,0.50,0.56}{##1}}}
\expandafter\def\csname PYG@tok@mf\endcsname{\def\PYG@tc##1{\textcolor[rgb]{0.13,0.50,0.31}{##1}}}
\expandafter\def\csname PYG@tok@err\endcsname{\def\PYG@bc##1{\setlength{\fboxsep}{0pt}\fcolorbox[rgb]{1.00,0.00,0.00}{1,1,1}{\strut ##1}}}
\expandafter\def\csname PYG@tok@mb\endcsname{\def\PYG@tc##1{\textcolor[rgb]{0.13,0.50,0.31}{##1}}}
\expandafter\def\csname PYG@tok@ss\endcsname{\def\PYG@tc##1{\textcolor[rgb]{0.32,0.47,0.09}{##1}}}
\expandafter\def\csname PYG@tok@sr\endcsname{\def\PYG@tc##1{\textcolor[rgb]{0.14,0.33,0.53}{##1}}}
\expandafter\def\csname PYG@tok@mo\endcsname{\def\PYG@tc##1{\textcolor[rgb]{0.13,0.50,0.31}{##1}}}
\expandafter\def\csname PYG@tok@kd\endcsname{\let\PYG@bf=\textbf\def\PYG@tc##1{\textcolor[rgb]{0.00,0.44,0.13}{##1}}}
\expandafter\def\csname PYG@tok@mi\endcsname{\def\PYG@tc##1{\textcolor[rgb]{0.13,0.50,0.31}{##1}}}
\expandafter\def\csname PYG@tok@kn\endcsname{\let\PYG@bf=\textbf\def\PYG@tc##1{\textcolor[rgb]{0.00,0.44,0.13}{##1}}}
\expandafter\def\csname PYG@tok@cpf\endcsname{\let\PYG@it=\textit\def\PYG@tc##1{\textcolor[rgb]{0.25,0.50,0.56}{##1}}}
\expandafter\def\csname PYG@tok@kr\endcsname{\let\PYG@bf=\textbf\def\PYG@tc##1{\textcolor[rgb]{0.00,0.44,0.13}{##1}}}
\expandafter\def\csname PYG@tok@s\endcsname{\def\PYG@tc##1{\textcolor[rgb]{0.25,0.44,0.63}{##1}}}
\expandafter\def\csname PYG@tok@kp\endcsname{\def\PYG@tc##1{\textcolor[rgb]{0.00,0.44,0.13}{##1}}}
\expandafter\def\csname PYG@tok@w\endcsname{\def\PYG@tc##1{\textcolor[rgb]{0.73,0.73,0.73}{##1}}}
\expandafter\def\csname PYG@tok@kt\endcsname{\def\PYG@tc##1{\textcolor[rgb]{0.56,0.13,0.00}{##1}}}
\expandafter\def\csname PYG@tok@sc\endcsname{\def\PYG@tc##1{\textcolor[rgb]{0.25,0.44,0.63}{##1}}}
\expandafter\def\csname PYG@tok@sb\endcsname{\def\PYG@tc##1{\textcolor[rgb]{0.25,0.44,0.63}{##1}}}
\expandafter\def\csname PYG@tok@k\endcsname{\let\PYG@bf=\textbf\def\PYG@tc##1{\textcolor[rgb]{0.00,0.44,0.13}{##1}}}
\expandafter\def\csname PYG@tok@se\endcsname{\let\PYG@bf=\textbf\def\PYG@tc##1{\textcolor[rgb]{0.25,0.44,0.63}{##1}}}
\expandafter\def\csname PYG@tok@sd\endcsname{\let\PYG@it=\textit\def\PYG@tc##1{\textcolor[rgb]{0.25,0.44,0.63}{##1}}}

\makeatother

\begin{document}

\maketitle
%\tableofcontents %git:rev-716-6e8165a
\phantomsection\label{index::doc}

\chapter{CHAPTERTITLE}
\label{index:chaptertitle}\label{index:coco-performance-assessment}% %\tableofcontents %git:rev-716-6e8165a is automatic with sphinx and moved behind the abstract
% by swap...py

\begin{abstract}
We present an any-time performance assessment for benchmarking numerical
optimization algorithms in a black-box scenario, applied within the \href{https://github.com/numbbo/coco}{COCO} benchmarking platform.
The performance assessment is based on \emph{runtimes} measured in number of objective function evaluations to reach one or several quality indicator target values.
We argue that runtime is the only available measure with a generic, meaningful, and quantitative interpretation.
We discuss the choice of the target values, runlength-based targets, and the aggregation of results by using simulated restarts, averages, and empirical distribution functions.
\end{abstract}\tableofcontents %git:rev-716-6e8165a
\newpage

\section{Introduction}
\label{index:introduction}
We present ideas and concepts for performance assessment when benchmarking numerical optimization algorithms in a black-box scenario.
Going beyond a simple ranking of algorithms, we aim
to provide a \emph{quantitative} and \emph{meaningful} performance assessment, which
allows for conclusions like \emph{algorithm A is seven times faster than algorithm
B} in solving a given problem or in solving problems with certain
characteristics.
For this end, we record algorithm \emph{runtimes, measured in
number of function evaluations} to reach predefined target values during the
algorithm run.

Runtimes represent the cost of optimization. Apart from a short, exploratory
experiment\footnote[1]{
The \href{https://github.com/numbbo/coco}{COCO} platform provides a CPU timing experiment to get a rough estimate of the time complexity of the algorithm \phantomsection\label{index:id8}{\hyperref[index:han2016ex]{\emph{{[}HAN2016ex{]}}}}.
}, we do not measure the algorithm cost in CPU or wall-clock time.
See for example \phantomsection\label{index:id5}{\hyperref[index:hoo1995]{\emph{{[}HOO1995{]}}}} for a discussion on shortcomings and
unfortunate consequences of benchmarking based on CPU time.

In the \href{https://github.com/numbbo/coco}{COCO} platform \phantomsection\label{index:id6}{\hyperref[index:han2016co]{\emph{{[}HAN2016co{]}}}}, we display average runtimes (aRT, see Section {\hyperref[index:averaging\string-runtime]{\emph{Averaging Runtime}}})
and the empirical distribution function of runtimes (ECDF, see Section {\hyperref[index:empirical\string-distribution\string-functions]{\emph{Empirical Distribution Functions}}}).
When displaying runtime distributions, we consider the aggregation over
target values and over subclasses of problems, or all problems.

\subsection{Terminology and Definitions}
\label{index:terminology-and-definitions}
In the \href{https://github.com/numbbo/coco}{COCO} framework in general, a \textbf{problem}, or problem instance triplet, \(p^3\), is defined by the search space dimension \(n\), the objective function \(f\), to be minimized, and its instance parameters \(\theta_i\) for instance \(i\).
More concisely, we consider a set of parametrized benchmark functions
\(f_\theta: \mathbb{R}^n \to \mathbb{R}^m, \theta \in \Theta\) and the corresponding problems \(p^3 = p(n, f_\theta, \theta_i)\).
Different instances vary by having different shifted optima, can use different rotations that are applied to the variables, have different optimal \(f\)-values, etc. \phantomsection\label{index:id9}{\hyperref[index:han2009fun]{\emph{{[}HAN2009fun{]}}}}.
The instance notion is introduced to generate repetition while avoiding possible exploitation of artificial function properties (like location of the optimum in zero).
The separation of dimension and instance parameters in the notation serves as a hint to indicate that we never aggregate over dimension and always aggregate over all \(\theta_i\)-values.

In the performance assessment setting, we associate to a problem instance
\(p^3\) a quality indicator mapping and a target value, such that a problem becomes a
quintuple \(p^5\).
Usually, the quality indicator remains the same for all problems, while we have
subsets of problems which only differ in their target value.
\begin{quote}
\end{quote}

\section{On Performance Measures}
\label{index:on-performance-measures}
Evaluating performance is necessarily based on performance \emph{measures}, the
definition of which plays a crucial role for the evaluation.
Here, we introduce a list of requirements a performance measure should satisfy in general, as well as in the context of black-box optimization specifically.
In general, a performance measure should be
\begin{itemize}
\item {} 
quantitative, as opposed to a simple \emph{ranking} of entrants (e.g., algorithms).
Ideally, the measure should be defined on a ratio scale (as opposed to an
interval or ordinal scale) \phantomsection\label{index:id10}{\hyperref[index:ste1946]{\emph{{[}STE1946{]}}}}, which allows to state that ``entrant A
is \(x\) \emph{times better} than entrant B''.\footnote[2]{
A variable which lives on a ratio scale has a meaningful zero,
allows for division, and can be taken to the logarithm in a meaningful way.
See for example \href{https://en.wikipedia.org/wiki/Level\_of\_measurement?oldid=478392481}{Level of measurement on Wikipedia}.
}

\item {} 
assuming a wide variation of values such that, for example, typical values do
not only range between 0.98 and 1.0,\footnote[3]{
A transformation like \(x\mapsto\log(1-x)\) could alleviate the problem
in this case, given it actually zooms in on relevant values.
}

\item {} 
well interpretable, in particular by having meaning and semantics attached to
the measured numbers,

\item {} 
relevant and meaningful with respect to the ``real world'',

\item {} 
as simple and as comprehensible as possible.

\end{itemize}

In the context of black-box optimization, the \textbf{runtime} to reach a target value, measured in number of function evaluations, satisfies all requirements.
Runtime is well-interpretable and meaningful with respect to the
real-world as it represents time needed to solve a problem.
Measuring number of function evaluations avoids the shortcomings of CPU
measurements that depend on parameters like the programming language, coding
style, machine used to run the experiment, etc., that are difficult or
impractical to control.
If however algorithm internal computations dominate wall-clock time in a
practical application, comparative runtime results \emph{in number of function
evaluations} can usually be adapted \emph{a posteri} to reflect the practical
scenario.
This hold also true for a speed up from parallelization.

\subsection{Quality Indicators}
\label{index:quality-indicators}\label{index:sec-verthori}
At each evaluation count (time step) \(t\) of an algorithm which optimizes a problem instance \(\theta_i\) of the function \(f_{\theta}\) in dimension \(n\), we apply a quality indicator mapping.
A quality indicator \(I\) maps the set of all solutions evaluated
so far (or recommended \phantomsection\label{index:id16}{\hyperref[index:han2016ex]{\emph{{[}HAN2016ex{]}}}}) to a problem-dependent real value.
Then, a runtime measurement can be obtained from each of a (large) set of
problem instances \(p^5 = p(n, f_\theta, \theta_i, I, I^\mathrm{target,
\theta_i}_{f})\).
The runtime on this problem instance is defined as the evaluation count
when the quality indicator value drops below the target for the first time, otherwise runtime remains undefined.

In the single-objective noiseless case, the quality indicator outputs
the best so far observed (i.e. minimal and feasible) function value.

In the single-objective noisy case, the quality indicator returns the 1\%-tile of
the function values of the last \(\lceil\ln(t + 3)^2 / 2\rceil\) evaluated
(or recommended) solutions.\footnote[4]{
This feature will only be available in the new implementation of the \href{https://github.com/numbbo/coco}{COCO} framework.
}

In the multi-objective case, the quality indicator is based on a negative
hypervolume indicator of the set of evaluated solutions (more specifically, the the non-dominated archive)
\phantomsection\label{index:id18}{\hyperref[index:bro2016]{\emph{{[}BRO2016{]}}}}, while other well- or lesser-known multi-objective quality indicators
are possible.

\subsection{Fixed-Budget versus Fixed-Target Approach}
\label{index:fixed-budget-versus-fixed-target-approach}
Starting from the most basic convergence graphs which plot the evolution of a
quality indicator, to be minimized, against the number of function evaluations,
there are essentially only two ways to measure the performance.
\begin{description}
\item[{fixed-budget approach:}] \leavevmode
We fix a maximal budget of function evaluations,
and measure the reached quality indicator value. A fixed search
budget can be pictured as drawing a \emph{vertical} line in the figure
(blue line in Figure {\hyperref[index:fig\string-horizontalvsvertical]{\emph{Fixed-Budget versus Fixed-Target}}}).

\item[{fixed-target approach:}] \leavevmode
We fix a target quality value and measure the number of function
evaluations, the \emph{runtime}, to reach this target. A fixed target can be
pictured as drawing a \emph{horizontal} line in the figure (red line in Figure
{\hyperref[index:fig\string-horizontalvsvertical]{\emph{Fixed-Budget versus Fixed-Target}}}).

\end{description}
\begin{figure}[htbp]
\centering
\capstart

\includegraphics[width=0.700\linewidth]{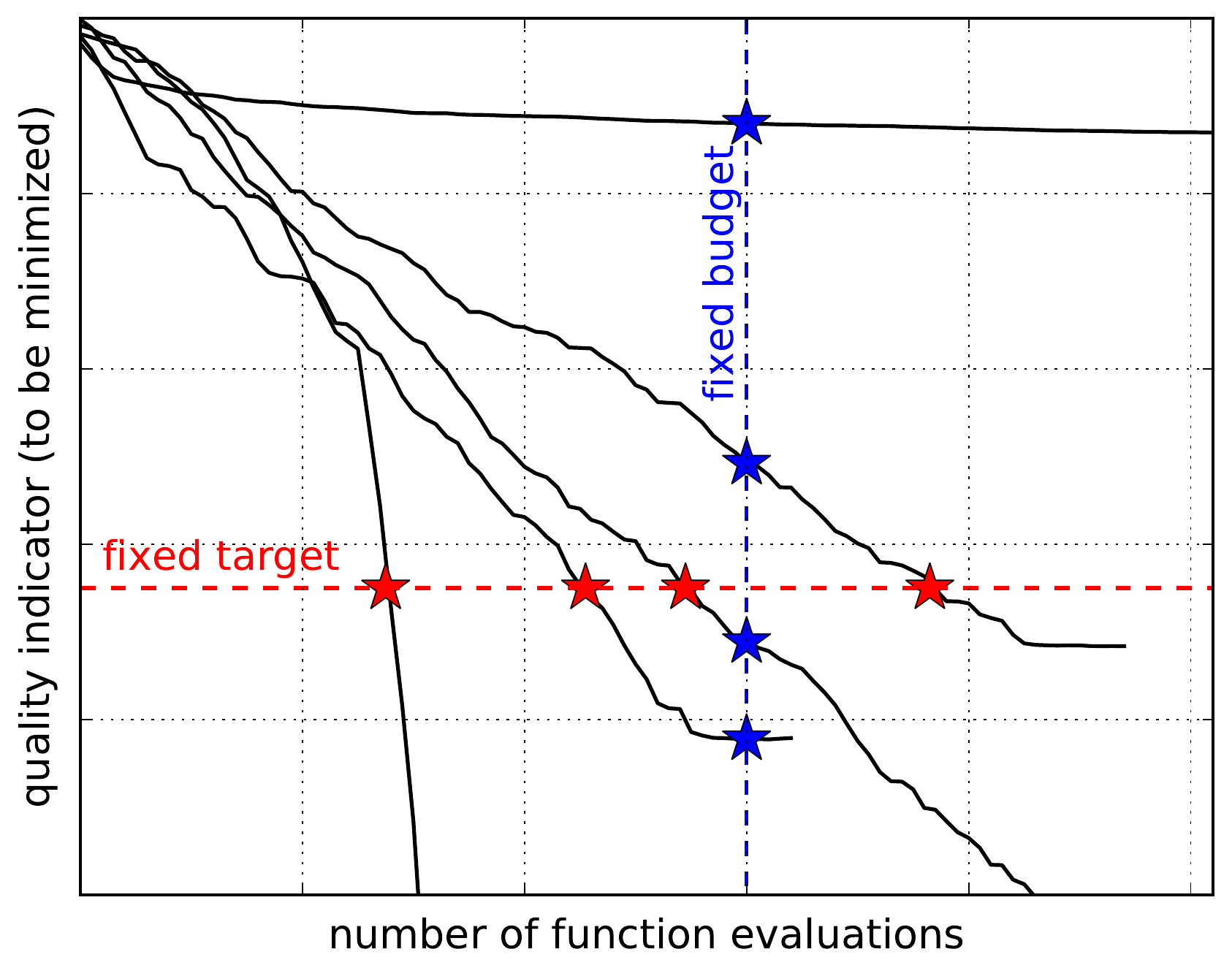}
\caption{\textbf{Fixed-Budget versus Fixed-Target}}{\small 
Illustration of fixed-budget view (vertical cuts) and fixed-target view
(horizontal cuts). Black lines depict the best quality indicator value
plotted versus number of function evaluations.
}\label{index:fig-horizontalvsvertical}\end{figure}

For the performance assessment of algorithms, the fixed-target approach is superior
to the fixed-budget approach since it gives \emph{quantitative and interpretable}
results.
\begin{itemize}
\item {} 
The fixed-budget approach (vertical cut) does not give \emph{quantitatively
interpretable}  data:
the observation that Algorithm A reaches a quality indicator value that is, say, two
times smaller than the one reached by Algorithm B has in general no
interpretable meaning, mainly because there is no \emph{a priori} way to determine
\emph{how much} more difficult it is to reach an indicator value that is two times
smaller.
This usually depends on the function, the definition of the
quality indicator and even the specific indicator values compared.

\item {} 
The fixed-target approach (horizontal cut)
\emph{measures the time} to
reach a target quality value. The measurement allows conclusions of the
type: Algorithm A is two (or ten, or a hundred) times faster than Algorithm B
in solving this problem.
The choice of the target value determines the difficulty and
often the characteristic of the problem to be solved.

\end{itemize}

Furthermore, for algorithms that are invariant under certain transformations
of the function value (for example under order-preserving transformations, as
comparison-based algorithms like DE, ES, PSO \phantomsection\label{index:id20}{\hyperref[index:aug2009]{\emph{{[}AUG2009{]}}}}), fixed-target measures are
invariant under these transformations if the target values are transformed accordingly. That is, only the horizontal line needs to be moved. Fixed-budget measures require the transformation of all resulting measurements individually.

\subsection{Missing Values}
\label{index:missing-values}
Investigating the Figure {\hyperref[index:fig\string-horizontalvsvertical]{\emph{Fixed-Budget versus Fixed-Target}}} more carefully, we find that not all graphs intersect with either the vertical or the horizontal line.
On the one hand, if the fixed budget is too large, the algorithm might solve the function before the budget is exceeded.\footnote[5]{
Even in continuous domain, from the view point of benchmarking,
or application in the real world, or numerical precision, the set of
solutions (or of solution sets) that indisputably solve the problem has a
volume larger than zero.
}
The algorithm performs better than the measurement is able to reflect, which can lead to a serious misinterpretations.
The remedy is to define a \emph{final} target value and measure the runtime if the final target is hit.\footnote[6]{
This is also advisable because declaring an algorithm better
when it reaches, say, \(\mathsf{const} + 10^{-30}\) instead of
\(\mathsf{const} + 10^{-10}\), is more often than not unjustified.
The former result may only indicate the lack of practical
termination conditions.
}

On the other hand, if the fixed target is too difficult, the algorithm may never hit the target under the given experimental conditions.\footnote[7]{
However, under mildly randomized conditions, for example with a randomized initial solution, the restarted algorithm reaches any attainable target with probability one. The time needed can of course well be beyond any reasonable practical limitations.
}
The algorithm performs worse than the experiment is able to reflect, while we still get a lower bound for this missing runtime instance.
A possible remedy is to run the algorithm longer.
Another possible remedy is to use the final quality indicator value as measurement.
This measurement however should only be interpreted as ranking result, defeating the original objective.
A third (impartial) remedy is to record the overall number of function evaluations of this run and use simulated restarts, see below.

\subsection{Target Value Setting}
\label{index:target-value-setting}
First, we define for each problem instance \(p^3 = (n, f_\theta, \theta_i)\)
a \emph{reference} quality indicator value, \(I^{\rm ref, \theta_i}\).
In the single-objective case this is the optimal function value.
In the multi-objective case this is the hypervolume indicator of an approximation of the Pareto front \phantomsection\label{index:id27}{\hyperref[index:bro2016]{\emph{{[}BRO2016{]}}}}.
Based on this reference value and a set of target \emph{precision} values, which are
independent of the instance \(\theta_i\), we define a target value
\begin{gather}
\begin{split}I^{\rm target,\theta_i} = I^{\rm ref,\theta_i} + \Delta I \enspace\end{split}\notag
\end{gather}
for each precision \(\Delta I\), giving rise to the product set of all problems \(p^3\) and all \(\Delta I\)-values.

\subsection{Runlength-based Target Values}
\label{index:runlength-based-target-values}
Runlength-based target values are a novel way to define the target values based
on a reference data set. Like for \emph{performance profiles} \phantomsection\label{index:id28}{\hyperref[index:dol2002]{\emph{{[}DOL2002{]}}}}, the
resulting empirical distribution can be interpreted \emph{relative to a reference
algorithm or a set of reference algorithms}.
Unlike for performance profiles, the resulting empirical distribution \emph{is} a
data profile \phantomsection\label{index:id29}{\hyperref[index:mor2009]{\emph{{[}MOR2009{]}}}} reflecting the true (opposed to relative) difficulty of the respective problems for the respective algorithm.

We assume to have given a reference data set with recorded runtimes to reach a
prescribed, usually large set of quality indicator target values\footnote[8]{
By default, the ratio between two neighboring \(\Delta I\) target precision values
is \(10^{0.2}\) and the largest \(\Delta I\) value is (dynamically) chosen such
that the first evaluation of the worst algorithm hits the target.
} as in the
fixed-target approach described above.
The reference data serve as a baseline upon which the runlength-based targets are  computed.
To simplify wordings we assume w.l.o.g. that a single reference \emph{algorithm} has generated this data set.

Now we choose a set of increasing reference \emph{budgets}. To each budget, starting with the smallest, we associate the easiest (largest) target for which (i) the average runtime (taken over all respective \(\theta_i\) instances, \(\mathrm{aRT}\), see below) of the reference algorithm \emph{exceeds} the budget and (ii, optionally) that had not been chosen for a smaller budget before. If such target does not exist, we take the final (smallest) target.

Like this, an algorithm that reaches a target within the associated budget is better than the reference algorithm on this problem.

Runlength-based targets are used in \href{https://github.com/numbbo/coco}{COCO} for the single-objective expensive optimization scenario.
The artificial best algorithm of BBOB-2009 (see below) is used as reference algorithm with either the five budgets of \(0.5n\), \(1.2n\), \(3n\), \(10n\), and \(50n\) function evaluations, where \(n\) is the problem
dimension, or with 31 targets evenly space on the log scale between \(0.5n\) and \(50n\) and without the optional constraint from (ii) above. In the latter case, the empirical distribution function of the runtimes of the reference algorithm shown in a \emph{semilogx} plot approximately resembles a diagonal straight line between the above two reference budgets.

Runlength-based targets have the \textbf{advantage} to make the target value setting less
dependent on the expertise of a human designer, because only the reference
\emph{budgets} have to be chosen a priori. Reference budgets, as runtimes, are
intuitively meaningful quantities, on which it is comparatively easy to decide
upon.
Runlength-based targets have the \textbf{disadvantage} to depend on the choice of a reference data set, that is, they depend on a set of reference algorithms.

\section{Runtime Computation}
\label{index:runtime-computation}
In the performance assessment context of \href{https://github.com/numbbo/coco}{COCO}, a problem instance can be defined by the quintuple search space dimension, function, instantiation parameters, quality indicator mapping, and quality indicator target value, \(p^5 = p(n, f_\theta, \theta_i, I, I^{{\rm target}, \theta_i})\).\footnote[9]{
From the definition of \(p\), we can generate a set of problems \(\mathcal{P}\) by varying one or several of the parameters. We never vary dimension \(n\) and always vary over all available instances \(\theta_i\) for generating \(\mathcal{P}.\)
}
For each benchmarked algorithm, a single runtime is measured on each problem instance.
From a \emph{single} run of the algorithm on the problem instance triple
\(p^3 = p(n, f_\theta, \theta_i)\), we obtain a runtime measurement for \emph{each} corresponding problem quintuple \(p^5\), more specifically, one for each target value which has been reached in this run, or equivalently, for each target precision.
This also reflects the anytime aspect of the performance evaluation in a single run.

Formally, the runtime \(\mathrm{RT}^{\rm s}(p)\) is a random variable that represents the number of function evaluations needed to reach the quality indicator target value for the first time.
A run or trial that reached the target value is called \emph{successful}.\footnote[10]{
The notion of success is directly linked to a target value. A run can be successful with respect to some target values (some problems) and unsuccessful with respect to others. Success also often refers to the final, most difficult, smallest target value, which implies success for all other targets.
}
For \emph{unsuccessful trials}, the runtime is not defined, but the overall number of function evaluations in the given trial is a random variable denoted by \(\mathrm{RT}^{\rm us}(p)\). For a single run, the value of \(\mathrm{RT}^{\rm us}(p)\) is the same for all failed targets.

We consider the conceptual \textbf{restart algorithm}.
Given an algorithm has a strictly positive probability \(p_{\mathrm{s}}\) to solve a
problem, independent restarts of the algorithm solve the problem with
probability one and exhibit the runtime
\phantomsection\label{index:equation-RTrestart}\begin{equation}
\label{index-RTrestart}
  % ":eq:`RTrestart`" becomes "\eqref{index-RTrestart}" in the LaTeX
\mathbf{RT}(n, f_\theta, \Delta I) = \sum_{j=1}^{J} \mathrm{RT}^{\rm us}_j(n,f_\theta,\Delta I) + \mathrm{RT}^{\rm s}(n,f_\theta,\Delta I)
\enspace,
\end{equation}
where \(J \sim \mathrm{BN}(1, 1 - p_{\rm s})\) is a random variable with negative binomial distribution that models the number of unsuccessful runs
until one success is observed and \(\mathrm{RT}^{\rm us}_j\) are independent
random variables corresponding to the evaluations in unsuccessful trials
\phantomsection\label{index:id34}{\hyperref[index:aug2005]{\emph{{[}AUG2005{]}}}}.
If the probability of success is one, \(J\) equals zero with probability one and the restart algorithm coincides with the original algorithm.

Generally, the above equation for \(\mathbf{RT}(n,f_\theta,\Delta I)\) expresses the runtime from repeated independent runs on the same problem instance (while the instance \(\theta_i\) is not given explicitly). For the performance evaluation in the \href{https://github.com/numbbo/coco}{COCO} framework, we apply the equation to runs on different instances \(\theta_i\), however instances from the same function, with the same dimension and the same target precision.

\subsection{Runs on Different Instances}
\label{index:runs-on-different-instances}
Different instantiations of the parametrized functions \(f_{\theta}\) are a natural way to represent randomized repetitions.
For example, different instances implement random translations of the search space and hence a translation of the optimum \phantomsection\label{index:id37}{\hyperref[index:han2009fun]{\emph{{[}HAN2009fun{]}}}}.
Randomized restarts on the other hand can be conducted from different initial points.
For translation invariant algorithms both mechanisms are equivalent and can be mutually exchanged.

We interpret thus runs performed on different instances \(\theta_1, \ldots, \theta_K\) as repetitions of the same problem.
Thereby we assume that instances of the same parametrized function \(f_{\theta}\) are
similar to each other, and more specifically that they exhibit the same runtime
distribution for each given \(\Delta I\).

We hence have for each parametrized problem a set of \(K\approx15\) independent runs, which are used to compute artificial runtimes of the conceptual restart algorithm.

\subsection{Simulated Restarts and Runtimes}
\label{index:simulated-restarts-and-runtimes}
The runtime of the conceptual restart algorithm as given in \eqref{index-RTrestart} is the basis for displaying performance within \href{https://github.com/numbbo/coco}{COCO}.
We use the \(K\) different runs on the same function and dimension to simulate virtual restarts with a fixed target precision.
We assume to have at least one successful run---otherwise, the runtime remains undefined, because the virtual procedure would never stop.
Then, we construct artificial, simulated runs from the available empirical data:
we repeatedly pick, uniformly at random with replacement, one of the \(K\) trials until we encounter a successful trial.
This procedure simulates a single sample of the virtually restarted algorithm from the given data.
As given in \eqref{index-RTrestart} as \(\mathbf{RT}(n,f_\theta,\Delta I)\), the measured, simulated runtime is the sum of the number of function evaluations from the unsuccessful trials added to the runtime of the last and successful trial.\footnote[11]{
In other words, we apply \eqref{index-RTrestart} such that \(\mathrm{RT}^{\mathrm{s}}\) is uniformly distributed over all measured runtimes from successful instances \(\theta_i\), \(\mathrm{RT}^{\mathrm{us}}\) is uniformly distributed over all evaluations seen in unsuccessful instances \(\theta_i\), and \(J\) has a negative binomial distribution \(\mathrm{BN}(1, q)\), where \(q\) is the number of unsuccessful instance divided by all instances.
}

\subsubsection{Bootstrapping Runtimes}
\label{index:bootstrapping-runtimes}
In practice, we repeat the above procedure between a hundred or even thousand times, thereby sampling \(N\) simulated runtimes from the same underlying distribution,
which then has striking similarities with the true distribution from a restarted algorithm \phantomsection\label{index:id40}{\hyperref[index:efr1994]{\emph{{[}EFR1994{]}}}}.
To reduce the variance in this procedure, when desired, the first trial in each sample is picked deterministically instead of randomly as the \(1 + (N~\mathrm{mod}~K)\)-th trial from the data.\footnote[12]{
The variance reducing effect is best exposed in the case where all runs are successful and \(N = K\), in which case each data is picked exactly once.
This example also suggests to apply a random permutation of the data before to simulate virtually restarted runs.
}
Picking the first trial data as specific instance \(\theta_i\) could also be
interpreted as applying simulated restarts to this specific instance rather than
to the entire set of problems \(\mathcal{P} = \{p(n, f_\theta, \theta_i, \Delta I) \;|\;
i=1,\dots,K\}\).

\subsubsection{Rationales and Limitations}
\label{index:rationales-and-limitations}
Simulated restarts aggregate some of the available data and thereby extend their range of interpretation.
\begin{itemize}
\item {} 
Simulated restarts allow in particular to compare algorithms with a wide range of different success probabilities by a single performance measure.\footnote[13]{
The range of success probabilities is bounded by the number of instances to roughly \(2/|K|.\)
} Conducting restarts is also valuable approach when addressing a difficult optimization problem in practice.

\item {} 
Simulated restarts rely on the assumption that the runtime distribution for each instance is the same. If this is not the case, they still provide a reasonable performance measure, however with less of a meaningful interpretation for the result.

\item {} 
The runtime of simulated restarts may heavily depend on \textbf{termination conditions} applied in the benchmarked algorithm, due to the evaluations spent in unsuccessful trials, compare \eqref{index-RTrestart}. This can be interpreted as disadvantage, when termination is considered as a trivial detail in the implementation---or as an advantage, when termination is considered a relevant component in the practical application of numerical optimization algorithms.

\item {} 
The maximal number of evaluations for which simulated runtimes are meaningful
and representative depends on the experimental conditions. If all runs are successful, no restarts are simulated and all runtimes are meaningful. If all runs terminated due to standard termination conditions in the used algorithm, simulated restarts reflect the original algorithm. However, if a maximal budget is imposed for the purpose of benchmarking, simulated restarts do not necessarily reflect the real performance. In this case and if the success probability drops below 1/2, the result is likely to give a too pessimistic viewpoint at or beyond the chosen maximal budget. See \phantomsection\label{index:id44}{\hyperref[index:han2016ex]{\emph{{[}HAN2016ex{]}}}} for a more in depth discussion on how to setup restarts in the experiments.

\item {} 
If only few or no successes have been observed, we can see large effects without statistical significance. Namely, 4/15 successes are not statistically significant against 0/15 successes on a 5\%-level.

\end{itemize}

\section{Averaging Runtime}
\label{index:sec-art}\label{index:averaging-runtime}
The average runtime (\(\mathrm{aRT}\)), introduced in \phantomsection\label{index:id46}{\hyperref[index:pri1997]{\emph{{[}PRI1997{]}}}} as ENES and
analyzed in \phantomsection\label{index:id47}{\hyperref[index:aug2005]{\emph{{[}AUG2005{]}}}} as success performance and referred to as
ERT in \phantomsection\label{index:id48}{\hyperref[index:han2009ex]{\emph{{[}HAN2009ex{]}}}}, estimates the expected runtime of the restart
algorithm given in \eqref{index-RTrestart}. Generally, the set of trials is
generated by varying \(\theta_i\) only.

We compute the \(\mathrm{aRT}\) from a set of trials as the sum of all evaluations in unsuccessful trials plus the sum of the runtimes in all successful trials, both divided by the number of successful trials.

\subsection{Motivation}
\label{index:motivation}
The expected runtime of the restart algorithm writes \phantomsection\label{index:id49}{\hyperref[index:aug2005]{\emph{{[}AUG2005{]}}}}
\begin{eqnarray*}
\mathbb{E}(\mathbf{RT}) & =
& \mathbb{E}(\mathrm{RT}^{\rm s})  + \frac{1-p_s}{p_s}
  \mathbb{E}(\mathrm{RT}^{\rm us})
\enspace,
\end{eqnarray*}
where \(p_{\mathrm{s}}\) is the probability of success of the algorithm and notations from above are used.

Given a data set with \(n_\mathrm{s}\ge1\) successful runs with runtimes \(\mathrm{RT}^{\rm s}_i\) and \(n_\mathrm{us}\) unsuccessful runs with \(\mathrm{RT}^{\rm us}_j\) evaluations, the average runtime reads
\begin{eqnarray*}
\mathrm{aRT}
& = &
\frac{1}{n_\mathrm{s}} \sum_i \mathrm{RT}^{\rm s}_i +
\frac{1-p_{\mathrm{s}}}{p_{\mathrm{s}}}\,
\frac{1}{n_\mathrm{us}} \sum_j \mathrm{RT}^{\rm us}_j
\\
& = &
\frac{\sum_i \mathrm{RT}^{\rm s}_i + \sum_j \mathrm{RT}^{\rm us}_j }{n_\mathrm{s}}
\\
& = &
\frac{\#\mathrm{FEs}}{n_\mathrm{s}}
\end{eqnarray*}
where \(p_{\mathrm{s}}\) is the fraction of successful trials, \(0/0\) is
understood as zero and \(\#\mathrm{FEs}\) is the number of function
evaluations conducted in all trials before to reach the given target precision.

\subsection{Rationale and Limitations}
\label{index:rationale-and-limitations}
The average runtime, \(\mathrm{aRT}\), is taken over different instances of the same function, dimension, and target precision, as these instances are interpreted as repetitions.
Taking the average is meaningful only if each instance obeys a similar distribution without heavy tail.
If one instance is considerably harder than the others, the average is dominated by this instance.
For this reason we do not average runtimes from different functions or different target precisions, which however could be done if the logarithm is taken first (geometric average).
Plotting the \(\mathrm{aRT}\) divided by dimension against dimension in a log-log plot is the recommended way to investigate the scaling behavior of an algorithm.

\section{Empirical Distribution Functions}
\label{index:empirical-distribution-functions}\label{index:sec-ecdf}
We display a set of simulated runtimes with the empirical cumulative
distribution function (ECDF), AKA empirical distribution function.
Informally, the ECDF displays the \emph{proportion of problems solved within a
specified budget}, where the budget is given on the \(x\)-axis.
More formally, an ECDF gives for each \(x\)-value the fraction of runtimes which do not exceed \(x\), where missing runtime values are counted in the denominator of the fraction.

\subsection{Rationale, Interpretation and Limitations}
\label{index:rationale-interpretation-and-limitations}
Empirical cumulative distribution functions are a universal way to display \emph{unlabeled} data in a condensed way without losing information.
They allow unconstrained aggregation, because each data point remains separately displayed, and they remain entirely meaningful under transformation of the data (e.g. taking the logarithm).
\begin{itemize}
\item {} 
The empirical distribution function from a set of problems where only the target value varies, recovers an upside-down convergence graph with the resolution steps defined by the targets \phantomsection\label{index:id50}{\hyperref[index:han2010]{\emph{{[}HAN2010{]}}}}.

\item {} 
When runs from several instances are aggregated, the association to the single run is lost, as is the association to the function when aggregating over several functions. This is particularly problematic for data from different dimensions, because dimension can be used as decision parameter for algorithm selection. Therefore, we do not aggregate over dimension.

\item {} 
The empirical distribution function can be read in two distinct ways.
\begin{description}
\item[{\(x\)-axis as independent variable:}] \leavevmode
for any budget (\(x\)-value),
we see the fraction of problems solved within the budget as \(y\)-value, where
the limit value to the right is the fraction of solved problems with the maximal
budget.

\item[{\(y\)-axis as independent variable:}] \leavevmode
for any fraction of easiest problems
(\(y\)-value), we see the maximal runtime observed on these problems on the
\(x\)-axis. When plotted in \emph{semilogx}, a horizontal shift indicates a runtime
difference by the respective factor, quantifiable, e.g., as ``five times
faster''. The area below the \(y\)-value and to the left of the graph reflects
the geometric runtime average on this subset of problems, the smaller the
better.

\end{description}

\end{itemize}

\subsection{Relation to Previous Work}
\label{index:relation-to-previous-work}
Empirical distribution functions over runtimes of optimization algorithms are also known as \emph{data profiles} \phantomsection\label{index:id51}{\hyperref[index:mor2009]{\emph{{[}MOR2009{]}}}}.
They are widely used for aggregating results from different functions and different dimensions to reach a single target precision \phantomsection\label{index:id52}{\hyperref[index:rio2012]{\emph{{[}RIO2012{]}}}}.
In the \href{https://github.com/numbbo/coco}{COCO} framework, we do not aggregation over dimension but aggregate often over a wide range of target precision values.

\subsection{Examples}
\label{index:examples}
We display in Figure {\hyperref[index:fig\string-ecdf]{\emph{ECDF}}} the ECDF of the (simulated) runtimes of
the pure random search algorithm on the set of problems formed by 15 instances of the sphere function (first function of the single-objective \code{bbob} test
suite) in dimension \(n=5\) each with 51 target precisions between \(10^2\) and \(10^{-8}\) uniform on a log-scale and 1000 bootstraps.
\begin{figure}[htbp]
\centering
\capstart

\includegraphics[width=0.700\linewidth]{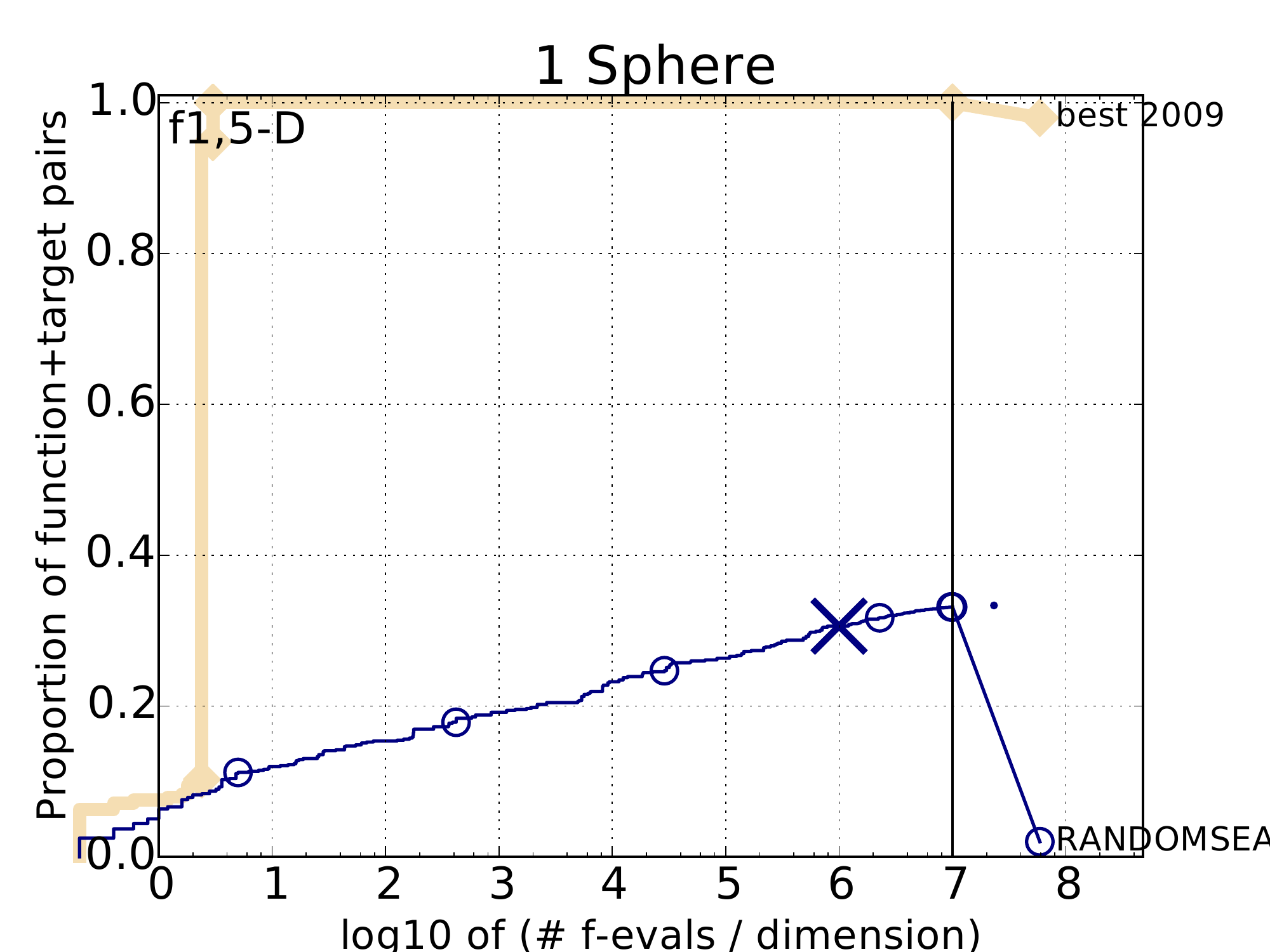}
\caption{ECDF}{\small 
Illustration of empirical (cumulative) distribution function (ECDF) of
runtimes on the sphere function using 51 relative targets uniform on a log
scale between \(10^2\) and \(10^{-8}\). The runtimes displayed
correspond to the pure random search algorithm in dimension 5. The cross on
the ECDF plots of \href{https://github.com/numbbo/coco}{COCO} represents the median of the maximal length of the
unsuccessful runs to solve the problems aggregated within the ECDF.
}\label{index:fig-ecdf}\end{figure}

We can see in this plot, for example, that almost 20 percent of the problems
were solved within \(10^3 \cdot n = 5 \cdot 10^3\) function evaluations.
Runtimes to the right of the cross at \(10^6\) have at least one unsuccessful run.
This can be concluded, because with pure random search each unsuccessful run exploits the maximum budget.
The small dot beyond \(x=10^7\) depicts the overall fraction of all successfully solved functions-target pairs, i.e., the fraction of \((f_\theta, \Delta I)\) pairs for which at least one trial (one \(\theta_i\) instantiation) was successful.

We usually divide the set of all (parametrized) benchmark
functions into subgroups sharing similar properties (for instance
separability, unimodality, ...) and display ECDFs which aggregate the
problems induced by these functions and all targets.
Figure {\hyperref[index:fig\string-ecdfgroup]{\emph{ECDF for a subgroup of functions}}} shows the result of random search on the first
five functions of the \emph{bbob} testsuite, separate (left) and aggregated (right).
\begin{figure}[htbp]
\centering
\capstart

\includegraphics[width=1.000\linewidth]{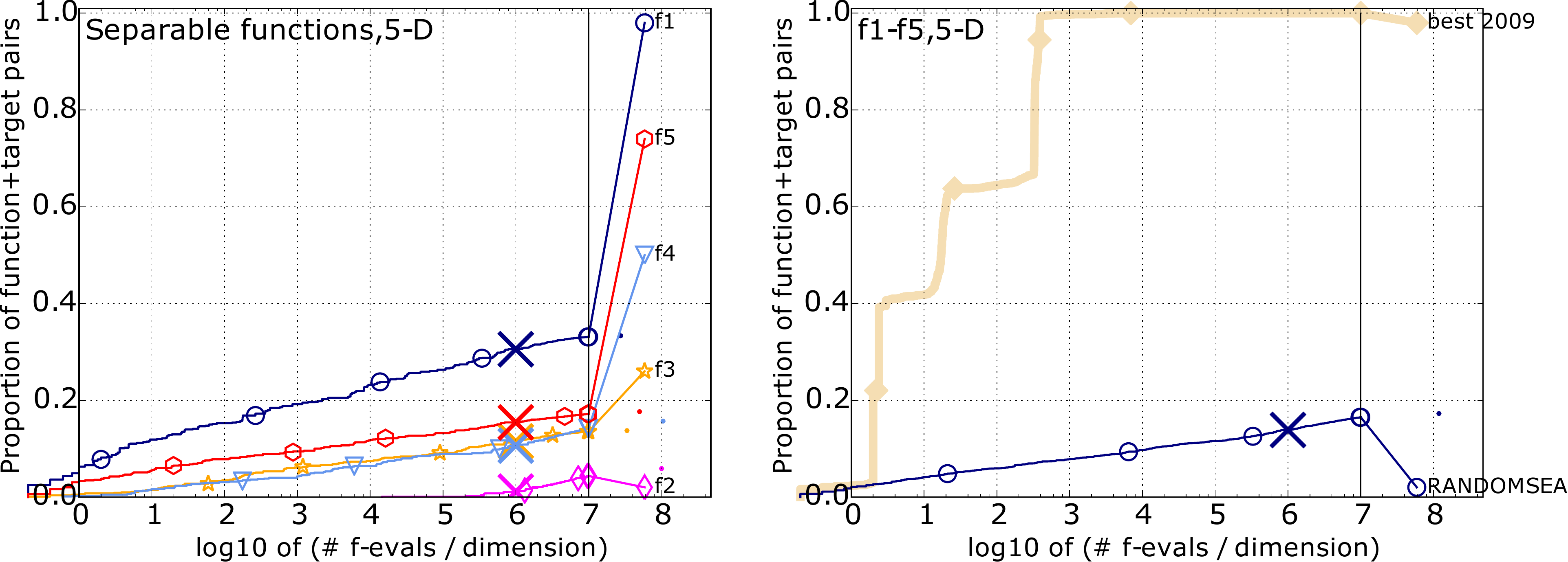}
\caption{ECDF for a subgroup of functions}{\small 
\textbf{Left:} ECDF of the runtime of the pure random search algorithm for
functions f1, f2, f3, f4 and f5 that constitute the group of
separable functions for the \code{bbob} testsuite over 51 target values.
\textbf{Right:} Aggregated ECDF of the same data, that is, all functions
in one graph.
}\label{index:fig-ecdfgroup}\end{figure}

Finally, we also naturally aggregate over all functions of the benchmark and
hence obtain one single ECDF per algorithm per dimension.
In Figure {\hyperref[index:fig\string-ecdfall]{\emph{ECDF over all functions and all targets}}}, the ECDF of different algorithms are displayed in
a single plot.
\begin{figure}[htbp]
\centering
\capstart

\includegraphics[width=1.000\linewidth]{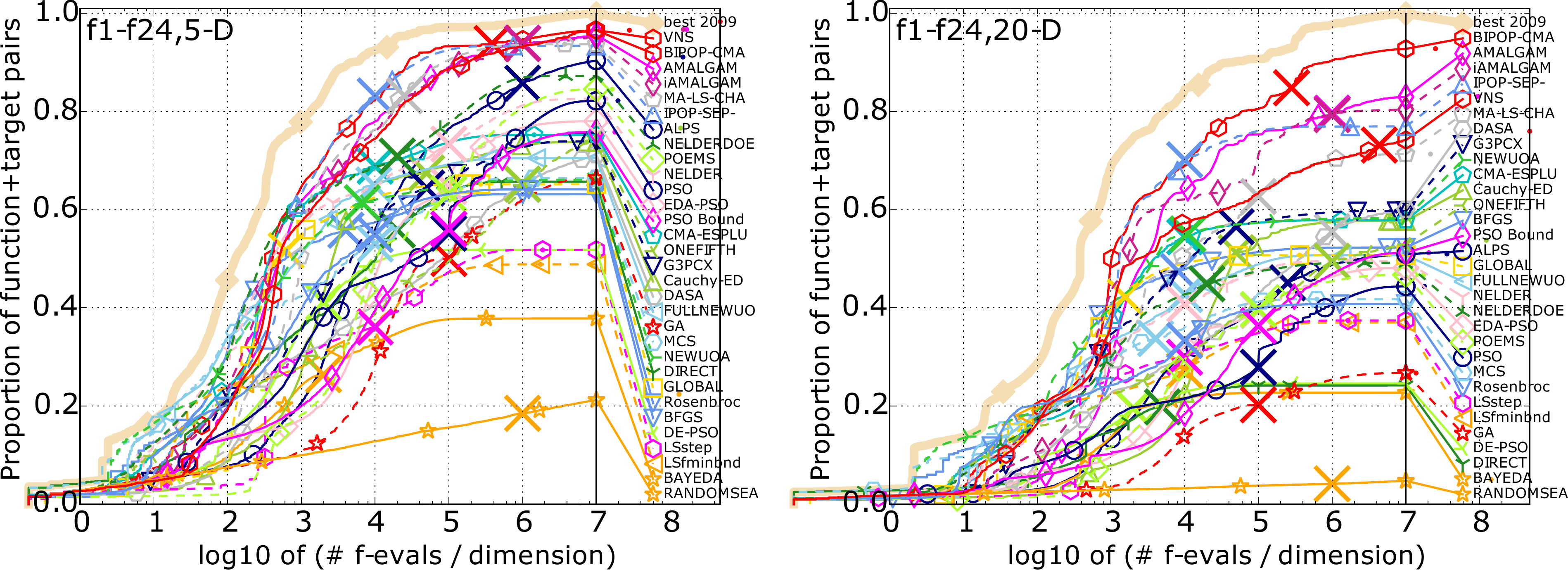}
\caption{ECDF over all functions and all targets}{\small 
ECDF of several algorithms benchmarked during the BBOB 2009 workshop
in dimension 5 (left) and in dimension 20 (right) when aggregating over all functions of the \code{bbob} suite.
}\label{index:fig-ecdfall}\end{figure}

The thick maroon line with diamond markers annotated as ``best 2009'' corresponds to the \textbf{artificial best 2009 algorithm}: for
each set of problems with the same function, dimension and target precision, we select the algorithm with the smallest \(\mathrm{aRT}\) from the \href{http://coco.gforge.inria.fr/doku.php?id=bbob-2009}{BBOB-2009 workshop} and use for these problems the data from the selected algorithm.
The algorithm is artificial because we may use even for different target values the runtime results from different algorithms.\footnote[14]{
The best 2009 curve is not guaranteed to be an upper
left envelope of the ECDF of all algorithms from which it is
constructed, that is, the ECDF of an algorithm from BBOB-2009 can
cross the best 2009 curve. This may typically happen if an algorithm
has for an easy target many short and few very
long runtimes such that its aRT is not the best but the short runtimes
show up to the left of the best 2009 graph.
}

We observe that the artificial best 2009 algorithm is about two to three time faster than the left envelope of all single algorithms and solves all problems in about \(10^7\, n\) function evaluations.
\section*{Acknowledgments}
This work was supported by the grant ANR-12-MONU-0009 (NumBBO)
of the French National Research Agency.

\renewcommand{\indexname}{Index}
\printindex
\end{document}